\pdfoutput=1
\documentclass[11pt]{article}

\usepackage{acl}

\usepackage{times}
\usepackage{amssymb}
\usepackage{pdfpages}
\usepackage{latexsym}
\usepackage{graphicx}
\usepackage{booktabs}
\usepackage{multirow}
\usepackage{caption}

\usepackage[T1]{fontenc}
\usepackage[utf8]{inputenc}

\usepackage{microtype}

\title{NU HLT at CMCL 2022 Shared Task: \\ Multilingual and Crosslingual Prediction of Human Reading Behavior in Universal Language Space}

\author{Joseph Marvin Imperial \\
Human Language Technology Lab (NU HLT)\\
  National University \\
  Manila, Philippines \\
  \texttt{jrimperial@national-u.edu.ph} \\}

\begin{document}
\maketitle
\begin{abstract}
In this paper, we present a unified model that works for both multilingual and crosslingual prediction of reading times of words in various languages. The secret behind the success of this model is in the preprocessing step where all words are transformed to their universal language representation via the International Phonetic Alphabet (IPA). To the best of our knowledge, this is the first study to favorably exploit this phonological property of language for the two tasks. Various feature types were extracted covering basic frequencies, n-grams, information theoretic, and psycholinguistically-motivated predictors for model training. A finetuned Random Forest model obtained best performance for both tasks with 3.8031 and 3.9065 MAE scores for mean first fixation duration (FFDAvg) and mean total reading time (TRTAvg) respectively\footnote{\url{https://github.com/imperialite/cmcl2022-unified-eye-tracking-ipa}}.
\end{abstract}

\section{Introduction}
Eye movement data has been one of the most used and most important resource that has pushed various interdisciplinary fields such as development studies, literacy, computer vision, and natural language processing research into greater heights. In a technical point of view, correctly determining theoretically grounded and cognitively plausible predictors of eye movement will allow opportunities to make computational systems leveraging on these properties to be more human-like \cite{sood2020improving}.

Common human reading prediction works make use of the standard Latin alphabet as it is internationally used. However, investigating eye movement and reading patterns in other non-Anglocentric writing systems such as Chinese and Bengali is as equally as important \cite{share2008anglocentricities, liversedge2016universality}. Fortunately, there is a growing number of previous works exploring multilinguality in eye tracking prediction both in data collection and novel prediction approaches. The study of \citet{liversedge2016universality} was the first to explore potential crosslinguality of Chinese, English and Finnish which differ in aspects of visual density, spacing, and orthography to name a few. The results of the study favorably support possible \textit{universality of representation} in reading. In the same vein, \citet{hollenstein-etal-2021-multilingual} was the first to try use of large finetuned multilingual language models like BERT \cite{devlin-etal-2019-bert} and XLM \cite{conneau2019cross} in a crosslingual setting to predict eye tracking features across English, Dutch, German, and Russian. Data-wise, the published works of \citet{siegelman2022expanding} for MECO, \citet{pynte2006influence} for the Dundee corpus, and \citet{cop2017presenting} for GECO have made significant impact in the field where they covered curation and collection of eye-tracking corpus for other languages in addition to English.

\section{Task Definition and Data}
The CMCL 2022 Shared Task \cite{hollenstein2022cmcl}\footnote{\url{https://cmclorg.github.io/shared\_task}} describes two challenges: predicting eye-tracking features in a \textbf{multilingual} and \textbf{crosslingual setup}. The eye movement dataset for this Shared Task contains sentences written in six languages: Mandarin Chinese \cite{pan2021beijing}, Hindi \cite{husain2015integration}, Russian \cite{laurinavichyute2019russian}, English \cite{luke2018provo, hollenstein2018zuco, hollenstein-etal-2020-zuco}, Dutch \cite{cop2017presenting}, and German \cite{jager2021potsdam}. The mean first fixation duration (FFDAvg) and mean total reading time (TRTAvg) as well as their corresponding standard deviations (FFDStd and TRTStd) are the four main eye-tracking features that need to be predicted by the participants through proposed computational means. For the multilingual task, the training, validation, and testing datasets conform to the identified six languages. While for the crosslingual task, a surprise language (Danish) is provided as the test dataset.

%%
%% FIGURE 1
%%

\begin{figure*}[!t]
\begin{center}

\includegraphics[width=0.50\textwidth, trim =3cm  0cm 3cm 0cm]{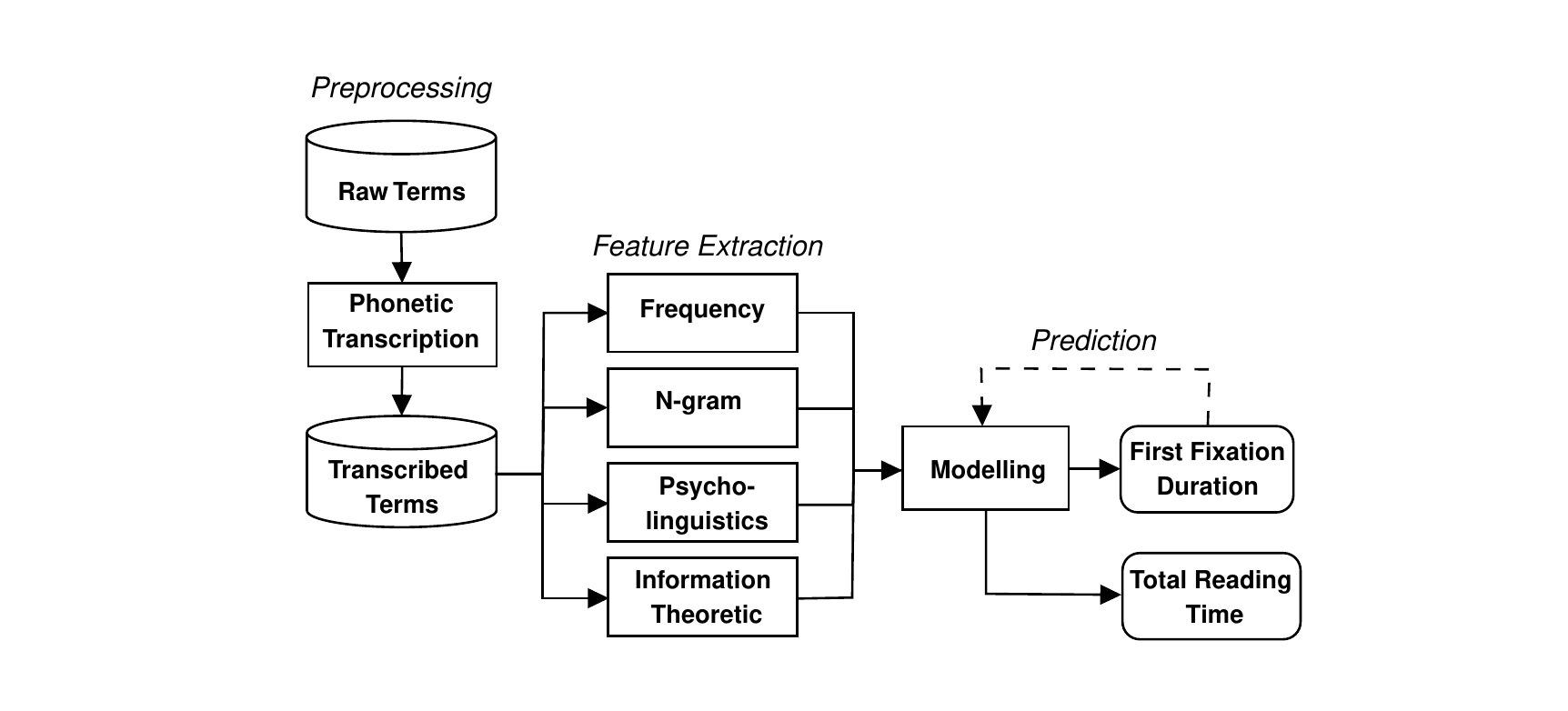}
\caption{The proposed \textbf{unified} approach to multilingual and crosslingual human reading pattern prediction in universal language space via IPA.}
\label{fig:methodology}
\end{center}
\end{figure*}

\section{Eye-Tracking Prediction in Universal Language Space}
The proposed solution in this work is inspired by both classical and recent previous works in speech recognition systems \cite{schultz1998multilingual, schultz2001language, dalmia2019phoneme} with multilingual and crosslingual capabilities through the transformation of words or similar sounding units in one global shared space using the International Phonetic Alphabet (IPA). This functionality allows models to generalize and adapt parameters to new languages while maintaining a stable vocabulary size for character representation. By definition, the IPA contains 107 characters for consonants and vowels, 31 for diacritics for modifying said consonants and vowels, and 17 signs to emphasize suprasegmental properties of phonemes such as stress and intonation \cite{international1999handbook}.

Figure~\ref{tab:mainResults} describes the unified methodology used for tackling both the multilinguality and crosslinguality challenge of the Shared Task. The backbone of this proposed solution lies with the phonetic transcription preprocessing step to convert the raw terms from the data written in Mandarin Chinese, Hindi, Russian, English, Dutch, and German to their IPA form. We used Epitran by \citet{mortensen2018epitran} for this process. The surprise language for the crosslingual task, Danish, is not currently supported by Epitran. We instead resorted to use Automatic Phonetic Transcriber\footnote{\url{http://tom.brondsted.dk/text2phoneme/}}, a paid transcription service that caters the Danish language. The transcription cost of the Danish test data is €15. 

\subsection{Feature Extraction}
After obtaining the phonetic transcriptions, a total of fourteen features based on various types were extracted spanning general frequencies, n-grams, based on information theory, and based on motivations from psycholinguistics.
\newline

\noindent\textbf{Frequency and Length Features}. The simplest features are frequency and length-based predictors. Studies have shown that the length of words correlate with fixation duration as long words would obviously take time to read \cite{rayner1977visual, hollenstein-beinborn-2021-relative}. For this study, we extracted the (a) word length (\texttt{word\_len}), (b) IPA length (\texttt{ipa\_len}), (c) IPA vowels count per term (\texttt{ipa\_count}), and (d) normalized IPA vowel count per term over length (\texttt{ipa\_norm}). 
\newline

\noindent\textbf{N-Gram Features}. Language model-based features is a classic in eye-tracking prediction research as they capture word probabilities through frequency. We extracted raw count of unique n-grams per word (\texttt{bigram\_count}, \texttt{trigram\_count}), raw count of total n-grams per term (\texttt{bigram\_sum}, \texttt{trigram\_sum}), and normalized counts over word length (\texttt{bigram\_norm}, \texttt{trigram\_norm}) for character bigrams and trigrams in IPA form guided by the general formula for n-gram modelling below:
\begin{equation}
    P(w_{n}\mid w_{n-N+1}^{n-1}) = \frac{C(w_{n-N+1}^{n-1}w_{n})}{C(w_{n-N+1}^{n-1})}
    %P(w_{i}^{n})= \prod_{k=1}^{n} P(w_{k}\mid w_{k-N+1}^{k-1})
\end{equation}

\noindent\textbf{Psycholinguistially-Motivated Features}. Features with theoretical grounding are more practical to use when invetigating phenomena in human reading. In line with this, we extracted two psycholinguistically-motivated features: \textbf{imageability} and \textbf{concreteness}. When reading, humans tend to visualize words and scenarios as they are formed in context. This measure of ease of how words or phrases can easily be visualized in the min from a verbal material is quantified as imageability \cite{lynch1964image, richardson1976imageability}. On the other hand, concreteness is a measure of lexical organization where words are easily perceived by the senses. In the example of \citet{schwanenflugel1988context}, words such as \textit{chair} or \textit{computer} are better understood than abstract words like \textit{freedom}. Words with high concreteness scores are better recalled from the mental lexicon than abstract words as they have better representation in the imaginal system \cite{altarriba1999concreteness}. We use these two features as we posit that the visualization and retrieval process of imageability and concreteness respectively can contribute to the reading time in milliseconds.

For this task, we used the crosslingual word embedding-based approximation for all the seven languages present in the dataset from the the work of \citet{ljubesic-etal-2018-predicting}.\newline

\noindent\textbf{Information Theoretic Features}.
Features inspired by information theory such as the concept of surprisal have thoroughly used in human reading pattern prediction \cite{hale2001probabilistic, levy2008expectation, demberg2008data, demberg2009computational, goodkind-bicknell-2018-predictive}. Surprisal describes that processing time of a word to be read is proportional to its negative log based on a probability given by context as shown below:
\begin{equation}
    \textrm{surprisal}(w_{i}) = -\textrm{log}_{2}\: P(w_{i}\mid w_{1}...w_{i-1})
\end{equation}

Thus, if a word is more likely to occur in its context, it is read more quickly \cite{shannon1948mathematical}. For this task, since words are converted to a universal language space, the correct terminology in this case is bits per phoneme or \textbf{phonotactic complexity} as coined by \citet{pimentel-etal-2020-phonotactic}.

While surprisal quantifies the word's predictability or processing cost during reading, we also obtain the \textbf{entropy} $H$ of each word $x$ from the corpus. The entropy quantifies the expected value of information from an event as shown in the formula below:
\begin{equation}
    H(X) = -\sum_{i=1}^{n}\:(\frac{count_{i}}{N})\:\textrm{log}_{2}\:(\frac{count_{i}}{N})
\end{equation}

where $count_{i}$ is the count of character $n_{i}$ and each word $N$ consists of $n$ characters. With this measure, a higher entropy score entails higher uncertainty for a word, thus, leading to increased reading time at the millisecond level.

\subsection{Model Training Setup}
We used four machine learning algorithms via WEKA \cite{witten2002data} for modelling the features with FFDAvg and TRTAvg: linear regression (\textbf{LinReg}), multilayer perceptron (\textbf{MLP}), random forest (\textbf{RF}), and k-Nearest Neighbors (\textbf{kNN}). We only used the finetuned RF model for the prediction of FFDAvg and TRTAvg. Meanwhile, FFDStd and TRTStd are obtained by using the top models of all the four algorithms, re-running them to get FFDAvg and TRTAvg, and calculating the standard deviation. For TRTAvg, we added the predicted FFDAvg from the best model as an additional feature as we posit that the first fixation duration is a contributor to the overall reading time.

%%
%% TABLE 1 - MAIN RESULTS
%%

\begin{table*}[!t]
\centering
\small
\begin{tabular}{@{}lcccc@{}}
\toprule
\multicolumn{1}{c}{\multirow{2}{*}{\bf Model}}    & \multicolumn{2}{c}{\bf FFDAvg}           & \multicolumn{2}{c}{\bf TRTAvg}          \\\cmidrule(lr){2-3}\cmidrule(lr){4-5}
\multicolumn{1}{c}{}                    & MAE             & RMSE            & MAE             & RMSE            \\

\midrule
\textbf{LinReg (k=10, M5)*$\dag$}                & \textbf{5.2361} & \textbf{6.7267} & \textbf{4.3419} & \textbf{7.0546} \\
LinReg (k=10, greedy)                & 5.2361          & 6.7267          & 4.3420          & 7.0545          \\
LinReg (k=10, none)                  & 5.2363          & 6.7274          & 4.3429          & 7.0594          \\

\midrule
\textbf{MLP (k=10, lr=0.005, m=0.2)*$\dag$} & \textbf{4.9898} & \textbf{6.4169} & \textbf{4.1744} & \textbf{6.2140} \\
MLP (k=10, lr=0.5, m=0.2)       & 6.7916          & 8.3791          & 4.8475          & 7.0840          \\
MLP (k=10, lr=0.005, m=0.002)   & 5.0018          & 6.4299          & 4.1862          & 6.2177          \\
MLP (k=10, lr=0.5, m=0.002)     & 6.4447          & 8.0110          & 4.9528          & 6.9668          \\
MLP (k=10, lr=0.0005, m=0.0002) & 5.5024          & 7.0474          & 4.2956          & 6.3823          \\

\midrule
\textbf{RF (k=10, iters = 100)*}        & \textbf{3.8031} & \textbf{5.2750}  & 3.9600          & 5.8446          \\
RF (k=10, iters = 100, 50\% feats)      & 3.8045          & 5.2766          & 3.9094          & 5.8015          \\
RF (k=10, iters = 100, 75\% feats$\dag$)      & 3.8056          & 5.2762          & \textbf{3.9065} & \textbf{5.8006} \\

\midrule
\textbf{kNN (k=10, nn=5, dist=euc)*}    & \textbf{4.3335} & \textbf{5.9651} & 4.2953          & 6.3741          \\
kNN (k=10, nn=10, dist=euc)             & 4.4263          & 6.0133          & 4.2053          & 6.2436          \\
kNN (k=10, nn=20, dist=euc)$\dag$             & 4.5646          & 6.1284          & \textbf{4.1793} & \textbf{6.2432}\\

\bottomrule
\end{tabular}
\caption{Results of predicting mean first fixation duration (FFDAvg) and mean total reading time (TRTAvg) using hyperparameter-tuned traditional supervised models. The tuned Random Forest (RF) model achieved the best performance which was used for both tasks of multilingual and crosslingual prediction. Top performing models from the four algorithm class were used for predicting the held-out test data to get the standard deviation of FFDAvg (*) and TRTAvg ($\dag$).}
\label{tab:mainResults}
\end{table*}

\section{Results}
Table~\ref{tab:mainResults} describes the main results of the experiments for predicting FFDAvg and TRTAvg using multiple finetuned supervised techniques evaluated through mean absolute error (MAE) and root mean squared error (RMSE). As mentioned previously, since the methodology used in this study cuts across multilingual and crosslingual tasks, the results reported in this applied are applicable to both. From the Table, the RF models outperformed the other three models in predicting FFDAVg and TRTAvg using 100\% and 75\% random selected features respectively and across 100 iterations. The RF model's effectivity can be attributed to its structure of multiple decision trees which normalize overfitting \cite{ho1995random}. Following RF in performance is kNN using Euclidean distance observing the same pattern as RF with different hyperparameter values such as 5 and 20 for the nearest neighbor for predicting FFDAvg and TRTAvg. On the other hand, both LinReg and MLP have no improvements regardless of hyperparameter values. For LinReg, using an M5 feature selection only provides extremely minor improvement in performances for FFDAvg and TRTAvg prediction. For MLP, using default values in WEKA for momentum and learning rate obtained the best performance similarly for for FFDAvg and TRTAvg prediction.

%%
%% TABLE 2 - CORRELATION
%%
\begin{table}[]
\centering
\small
\begin{tabular}{lr|lr}
\toprule
\multicolumn{2}{c|}{\bf FFDAvg}                  & \multicolumn{2}{c}{\bf TRTAvg}                   \\ \midrule
\multicolumn{1}{l}{bigram\_norm}  & -0.1751 & \multicolumn{1}{l}{FFDAvg}         & 0.8068  \\ 
\multicolumn{1}{l}{trigram\_norm} & -0.1393 & \multicolumn{1}{l}{bigram\_count}  & 0.2219  \\ 
\multicolumn{1}{l}{word\_len}     & -0.1334 & \multicolumn{1}{l}{trigram\_count} & 0.2156  \\ 
\multicolumn{1}{l}{bigram\_sum}   & -0.1304 & \multicolumn{1}{l}{phonetic\_comp} & -0.2107 \\ 
\multicolumn{1}{l}{trigram\_sum}  & -0.1101 & \multicolumn{1}{l}{ipa\_ent}       & 0.1925  \\ 
\multicolumn{1}{l}{imageability}  & 0.1101  & \multicolumn{1}{l}{ipa\_len}       & 0.1921  \\ 
\multicolumn{1}{l}{concreteness} & 0.1044 & \multicolumn{1}{l}{trigram\_norm} & \multicolumn{1}{l}{-0.1886}  \\ 

\bottomrule
\end{tabular}
\caption{Top 7 predictors for FFDAvg and TRTAvg with the highest correlation coefficients. }
\label{tab:correlation}
\end{table}

\subsection{Feature Importance}

Viewing the results in a correlation analysis perspective, Table~\ref{tab:correlation} shows the top 50\% of the predictors, total 7, which are significantly correlated with FFDAvg and TRTAvg respectively. Only one predictor is common for both values, the normalized trigrams in IPA space which is fairly high in FFDAvg along with normalized bigrams than in TRTAvg. This may hint that normalized n-gram features may be plausible features of eye movement only for first passes over the word and not with the total accumulated time of fixations. Likewise, the psycholinguistically-motivated features, imageability and concreteness, were only seen in the FFDAvg section as well proving their potential plausibility for the same observation. All the length-based features such as word, IPA, bigram, and trigram-based counts were considered as top predictors for FFDAvg and TRTAvg. This unsurprisingly supports the results from the classical work of \citet{rayner1977visual} on correlation of lengths with fixations. Lastly, the strong correlation of first fixation duration with the total reading time with a score of $r$ = 0.8068 proves the theoretical grounding of the proposed methodology as stated in Figure~\ref{fig:methodology} albeit in post-hoc.

\section{Conclusion}
Precise eye movement datasets in multiple languages are considered one of the most important contributions that benefit various interdisciplinary fields such as psycholinguistics, developmental studies, behavioral studies, computer vision, and natural language processing. In this paper, we present a novel method of transforming multilingual eye-tracking data (English, Mandarin, Hindi, Russian, German, Dutch, and Danish) to their IPA equivalent, enforcing a single vocabulary space which allows competitive results for both multilingual and crosslingual tasks in a regression analysis setup. Future directions of this paper can explore more cognitively and theoretically plausible features that can be extracted as well as deeper interpretation studies of the predictive models trained.

%\section*{Acknowledgements}

\bibliography{anthology,references}

\begin{thebibliography}{38}
\expandafter\ifx\csname natexlab\endcsname\relax\def\natexlab#1{#1}\fi

\bibitem[{Altarriba et~al.(1999)Altarriba, Bauer, and
  Benvenuto}]{altarriba1999concreteness}
Jeanette Altarriba, Lisa~M Bauer, and Claudia Benvenuto. 1999.
\newblock Concreteness, context availability, and imageability ratings and word
  associations for abstract, concrete, and emotion words.
\newblock \emph{Behavior Research Methods, Instruments, \& Computers},
  31(4):578--602.

\bibitem[{Association et~al.(1999)Association, Staff
  et~al.}]{international1999handbook}
International~Phonetic Association, International Phonetic~Association Staff,
  et~al. 1999.
\newblock \emph{Handbook of the International Phonetic Association: A guide to
  the use of the International Phonetic Alphabet}.
\newblock Cambridge University Press.

\bibitem[{Conneau and Lample(2019)}]{conneau2019cross}
Alexis Conneau and Guillaume Lample. 2019.
\newblock Cross-lingual language model pretraining.
\newblock \emph{Advances in Neural Information Processing Systems}, 32.

\bibitem[{Cop et~al.(2017)Cop, Dirix, Drieghe, and Duyck}]{cop2017presenting}
Uschi Cop, Nicolas Dirix, Denis Drieghe, and Wouter Duyck. 2017.
\newblock Presenting geco: An eyetracking corpus of monolingual and bilingual
  sentence reading.
\newblock \emph{Behavior Research Methods}, 49(2):602--615.

\bibitem[{Dalmia et~al.(2019)Dalmia, Li, Black, and Metze}]{dalmia2019phoneme}
Siddharth Dalmia, Xinjian Li, Alan~W Black, and Florian Metze. 2019.
\newblock Phoneme level language models for sequence based low resource asr.
\newblock In \emph{ICASSP 2019-2019 IEEE International Conference on Acoustics,
  Speech and Signal Processing (ICASSP)}, pages 6091--6095. IEEE.

\bibitem[{Demberg and Keller(2008)}]{demberg2008data}
Vera Demberg and Frank Keller. 2008.
\newblock Data from eye-tracking corpora as evidence for theories of syntactic
  processing complexity.
\newblock \emph{Cognition}, 109(2):193--210.

\bibitem[{Demberg and Keller(2009)}]{demberg2009computational}
Vera Demberg and Frank Keller. 2009.
\newblock A computational model of prediction in human parsing: Unifying
  locality and surprisal effects.
\newblock In \emph{Proceedings of the Annual Meeting of the Cognitive Science
  Society}, volume~31.

\bibitem[{Devlin et~al.(2019)Devlin, Chang, Lee, and
  Toutanova}]{devlin-etal-2019-bert}
Jacob Devlin, Ming-Wei Chang, Kenton Lee, and Kristina Toutanova. 2019.
\newblock \href {https://doi.org/10.18653/v1/N19-1423} {{BERT}: Pre-training of
  deep bidirectional transformers for language understanding}.
\newblock In \emph{Proceedings of the 2019 Conference of the North {A}merican
  Chapter of the Association for Computational Linguistics: Human Language
  Technologies, Volume 1 (Long and Short Papers)}, pages 4171--4186,
  Minneapolis, Minnesota. Association for Computational Linguistics.

\bibitem[{Goodkind and Bicknell(2018)}]{goodkind-bicknell-2018-predictive}
Adam Goodkind and Klinton Bicknell. 2018.
\newblock \href {https://doi.org/10.18653/v1/W18-0102} {Predictive power of
  word surprisal for reading times is a linear function of language model
  quality}.
\newblock In \emph{Proceedings of the 8th Workshop on Cognitive Modeling and
  Computational Linguistics ({CMCL} 2018)}, pages 10--18, Salt Lake City, Utah.
  Association for Computational Linguistics.

\bibitem[{Hale(2001)}]{hale2001probabilistic}
John Hale. 2001.
\newblock A probabilistic earley parser as a psycholinguistic model.
\newblock In \emph{Second Meeting of the North American Chapter of the
  Association for Computational Linguistics}.

\bibitem[{Ho(1995)}]{ho1995random}
Tin~Kam Ho. 1995.
\newblock Random decision forests.
\newblock In \emph{Proceedings of 3rd International Conference on Document
  Analysis and Recognition}, volume~1, pages 278--282. IEEE.

\bibitem[{Hollenstein and Beinborn(2021)}]{hollenstein-beinborn-2021-relative}
Nora Hollenstein and Lisa Beinborn. 2021.
\newblock \href {https://doi.org/10.18653/v1/2021.acl-short.19} {Relative
  importance in sentence processing}.
\newblock In \emph{Proceedings of the 59th Annual Meeting of the Association
  for Computational Linguistics and the 11th International Joint Conference on
  Natural Language Processing (Volume 2: Short Papers)}, pages 141--150,
  Online. Association for Computational Linguistics.

\bibitem[{Hollenstein et~al.(2022)Hollenstein, Chersoni, Jacobs, Oseki,
  Prévot, and Santus}]{hollenstein2022cmcl}
Nora Hollenstein, Emmanuel Chersoni, Cassandra Jacobs, Yohei Oseki, Laurent
  Prévot, and Enrico Santus. 2022.
\newblock {CMCL} 2022 {S}hared {T}ask on {M}ultilingual and {C}rosslingual
  {P}rediction of {H}uman {R}eading {B}ehavior.
\newblock In \emph{Proceedings of the Workshop on Cognitive Modeling and
  Computational Linguistics}.

\bibitem[{Hollenstein et~al.(2021)Hollenstein, Pirovano, Zhang, J{\"a}ger, and
  Beinborn}]{hollenstein-etal-2021-multilingual}
Nora Hollenstein, Federico Pirovano, Ce~Zhang, Lena J{\"a}ger, and Lisa
  Beinborn. 2021.
\newblock \href {https://doi.org/10.18653/v1/2021.naacl-main.10} {Multilingual
  language models predict human reading behavior}.
\newblock In \emph{Proceedings of the 2021 Conference of the North American
  Chapter of the Association for Computational Linguistics: Human Language
  Technologies}, pages 106--123, Online. Association for Computational
  Linguistics.

\bibitem[{Hollenstein et~al.(2018)Hollenstein, Rotsztejn, Troendle, Pedroni,
  Zhang, and Langer}]{hollenstein2018zuco}
Nora Hollenstein, Jonathan Rotsztejn, Marius Troendle, Andreas Pedroni,
  Ce~Zhang, and Nicolas Langer. 2018.
\newblock Zu{C}o, a simultaneous {EEG} and eye-tracking resource for natural
  sentence reading.
\newblock \emph{Scientific Data}, 5(1):1--13.

\bibitem[{Hollenstein et~al.(2020)Hollenstein, Troendle, Zhang, and
  Langer}]{hollenstein-etal-2020-zuco}
Nora Hollenstein, Marius Troendle, Ce~Zhang, and Nicolas Langer. 2020.
\newblock \href {https://aclanthology.org/2020.lrec-1.18} {{Z}u{C}o 2.0: A
  dataset of physiological recordings during natural reading and annotation}.
\newblock In \emph{Proceedings of the 12th Language Resources and Evaluation
  Conference}, pages 138--146, Marseille, France. European Language Resources
  Association.

\bibitem[{Husain et~al.(2015)Husain, Vasishth, and
  Srinivasan}]{husain2015integration}
Samar Husain, Shravan Vasishth, and Narayanan Srinivasan. 2015.
\newblock Integration and prediction difficulty in hindi sentence
  comprehension: Evidence from an eye-tracking corpus.
\newblock \emph{Journal of Eye Movement Research}, 8(2).

\bibitem[{J{\"a}ger et~al.(2021)J{\"a}ger, Kern, and Haller}]{jager2021potsdam}
Lena~A J{\"a}ger, Thomas Kern, and Patrick Haller. 2021.
\newblock Potsdam {T}extbook {C}orpus (potec).

\bibitem[{Laurinavichyute et~al.(2019)Laurinavichyute, Sekerina, Alexeeva,
  Bagdasaryan, and Kliegl}]{laurinavichyute2019russian}
Anna~K Laurinavichyute, Irina~A Sekerina, Svetlana Alexeeva, Kristine
  Bagdasaryan, and Reinhold Kliegl. 2019.
\newblock Russian sentence corpus: Benchmark measures of eye movements in
  reading in russian.
\newblock \emph{Behavior Research Methods}, 51(3):1161--1178.

\bibitem[{Levy(2008)}]{levy2008expectation}
Roger Levy. 2008.
\newblock Expectation-based syntactic comprehension.
\newblock \emph{Cognition}, 106(3):1126--1177.

\bibitem[{Liversedge et~al.(2016)Liversedge, Drieghe, Li, Yan, Bai, and
  Hy{\"o}n{\"a}}]{liversedge2016universality}
Simon~P Liversedge, Denis Drieghe, Xin Li, Guoli Yan, Xuejun Bai, and Jukka
  Hy{\"o}n{\"a}. 2016.
\newblock Universality in eye movements and reading: A trilingual
  investigation.
\newblock \emph{Cognition}, 147:1--20.

\bibitem[{Ljube{\v{s}}i{\'c} et~al.(2018)Ljube{\v{s}}i{\'c}, Fi{\v{s}}er, and
  Peti-Stanti{\'c}}]{ljubesic-etal-2018-predicting}
Nikola Ljube{\v{s}}i{\'c}, Darja Fi{\v{s}}er, and Anita Peti-Stanti{\'c}. 2018.
\newblock \href {https://doi.org/10.18653/v1/W18-3028} {Predicting concreteness
  and imageability of words within and across languages via word embeddings}.
\newblock In \emph{Proceedings of The Third Workshop on Representation Learning
  for {NLP}}, pages 217--222, Melbourne, Australia. Association for
  Computational Linguistics.

\bibitem[{Luke and Christianson(2018)}]{luke2018provo}
Steven~G Luke and Kiel Christianson. 2018.
\newblock The provo corpus: A large eye-tracking corpus with predictability
  norms.
\newblock \emph{Behavior Research Methods}, 50(2):826--833.

\bibitem[{Lynch(1964)}]{lynch1964image}
Kevin Lynch. 1964.
\newblock \emph{The image of the city}.
\newblock MIT press.

\bibitem[{Mortensen et~al.(2018)Mortensen, Dalmia, and
  Littell}]{mortensen2018epitran}
David~R Mortensen, Siddharth Dalmia, and Patrick Littell. 2018.
\newblock Epitran: Precision g2p for many languages.
\newblock In \emph{Proceedings of the Eleventh International Conference on
  Language Resources and Evaluation (LREC 2018)}.

\bibitem[{Pan et~al.(2021)Pan, Yan, Richter, Shu, and Kliegl}]{pan2021beijing}
Jinger Pan, Ming Yan, Eike~M Richter, Hua Shu, and Reinhold Kliegl. 2021.
\newblock The beijing sentence corpus: A chinese sentence corpus with eye
  movement data and predictability norms.
\newblock \emph{Behavior Research Methods}, pages 1--12.

\bibitem[{Pimentel et~al.(2020)Pimentel, Roark, and
  Cotterell}]{pimentel-etal-2020-phonotactic}
Tiago Pimentel, Brian Roark, and Ryan Cotterell. 2020.
\newblock \href {https://doi.org/10.1162/tacl_a_00296} {Phonotactic complexity
  and its trade-offs}.
\newblock \emph{Transactions of the Association for Computational Linguistics},
  8:1--18.

\bibitem[{Pynte and Kennedy(2006)}]{pynte2006influence}
Joel Pynte and Alan Kennedy. 2006.
\newblock An influence over eye movements in reading exerted from beyond the
  level of the word: Evidence from reading english and french.
\newblock \emph{Vision Research}, 46(22):3786--3801.

\bibitem[{Rayner(1977)}]{rayner1977visual}
Keith Rayner. 1977.
\newblock Visual attention in reading: Eye movements reflect cognitive
  processes.
\newblock \emph{Memory \& cognition}, 5(4):443--448.

\bibitem[{Richardson(1976)}]{richardson1976imageability}
John~TE Richardson. 1976.
\newblock Imageability and concreteness.
\newblock \emph{Bulletin of the Psychonomic Society}, 7(5):429--431.

\bibitem[{Schultz and Waibel(1998)}]{schultz1998multilingual}
Tanja Schultz and Alex Waibel. 1998.
\newblock Multilingual and crosslingual speech recognition.
\newblock In \emph{Proc. DARPA Workshop on Broadcast News Transcription and
  Understanding}, pages 259--262. Citeseer.

\bibitem[{Schultz and Waibel(2001)}]{schultz2001language}
Tanja Schultz and Alex Waibel. 2001.
\newblock Language-independent and language-adaptive acoustic modeling for
  speech recognition.
\newblock \emph{Speech Communication}, 35(1-2):31--51.

\bibitem[{Schwanenflugel et~al.(1988)Schwanenflugel, Harnishfeger, and
  Stowe}]{schwanenflugel1988context}
Paula~J Schwanenflugel, Katherine~Kip Harnishfeger, and Randall~W Stowe. 1988.
\newblock Context availability and lexical decisions for abstract and concrete
  words.
\newblock \emph{Journal of Memory and Language}, 27(5):499--520.

\bibitem[{Shannon(1948)}]{shannon1948mathematical}
Claude~Elwood Shannon. 1948.
\newblock A mathematical theory of communication.
\newblock \emph{The Bell system technical journal}, 27(3):379--423.

\bibitem[{Share(2008)}]{share2008anglocentricities}
David~L Share. 2008.
\newblock On the anglocentricities of current reading research and practice:
  the perils of overreliance on an" outlier" orthography.
\newblock \emph{Psychological Bulletin}, 134(4):584.

\bibitem[{Siegelman et~al.(2022)Siegelman, Schroeder, Acart{\"u}rk, Ahn,
  Alexeeva, Amenta, Bertram, Bonandrini, Brysbaert, Chernova
  et~al.}]{siegelman2022expanding}
Noam Siegelman, Sascha Schroeder, Cengiz Acart{\"u}rk, Hee-Don Ahn, Svetlana
  Alexeeva, Simona Amenta, Raymond Bertram, Rolando Bonandrini, Marc Brysbaert,
  Daria Chernova, et~al. 2022.
\newblock Expanding horizons of cross-linguistic research on reading: The
  multilingual eye-movement corpus (meco).
\newblock \emph{Behavior Research Methods}, pages 1--21.

\bibitem[{Sood et~al.(2020)Sood, Tannert, M{\"u}ller, and
  Bulling}]{sood2020improving}
Ekta Sood, Simon Tannert, Philipp M{\"u}ller, and Andreas Bulling. 2020.
\newblock Improving natural language processing tasks with human gaze-guided
  neural attention.
\newblock \emph{Advances in Neural Information Processing Systems},
  33:6327--6341.

\bibitem[{Witten and Frank(2002)}]{witten2002data}
Ian~H Witten and Eibe Frank. 2002.
\newblock Data mining: practical machine learning tools and techniques with
  java implementations.
\newblock \emph{Acm Sigmod Record}, 31(1):76--77.

\end{thebibliography}
\bibliographystyle{acl_natbib}

%\appendix

%\section{Example Appendix}
%\label{sec:appendix}

%This is an appendix.

\end{document}